\documentclass[letterpaper, 10 pt, journal, twoside]{IEEEtran}

\usepackage{amsmath}
\usepackage{mathtools}
\usepackage[linesnumbered,ruled,vlined]{algorithm2e}
\usepackage{amsfonts}
\usepackage{xr-hyper}
\usepackage{hyperref}
\usepackage{amssymb}
\usepackage{makecell}
\usepackage{booktabs,tabularx,colortbl,multirow,array,makecell}
\usepackage{graphics} 
\usepackage{epsfig} 
\usepackage{mathptmx} 
\usepackage{times} 
\usepackage[linesnumbered,ruled,vlined]{algorithm2e}
\usepackage{subcaption}
\usepackage[utf8]{inputenc}
\usepackage{graphicx}
\usepackage{bm}
\usepackage{xcolor}
\usepackage{enumitem}
\usepackage{cleveref}
\usepackage{tabularx}
\usepackage{pifont}
\newcommand{\cmark}{\ding{51}}%
\newcommand{\xmark}{\ding{55}}%

\setlist{nolistsep}

\DeclareMathOperator*{\argmax}{arg\,max}

\newcommand{\benchmark}{\texttt{LEMMA}\xspace}

\title{LEMMA: Learning Language-Conditioned Multi-Robot Manipulation}

\author{
Ran Gong$^{1}$, Xiaofeng Gao$^{2}$, Qiaozi Gao$^{2}$, Suhaila Shakiah$^{2}$, Govind Thattai$^{2}$, Gaurav S. Sukhatme$^{2,3}$
\thanks{Manuscript received: April, 18, 2023; Revised August, 1, 2023; Accepted August, 21, 2023.}
\thanks{This paper was recommended for publication by Editor Aleksandra Faust upon evaluation of the Associate Editor and Reviewers' comments. This work was supported by Amazon Alexa AI. Corresponding author: Xiaofeng Gao}
\thanks{Project website: \href{https://lemma-benchmark.github.io}{https://lemma-benchmark.github.io}}
\thanks{$^{1}$Center for Vision, Cognition, Learning, and Autonomy, UCLA. Email: nikepupu@ucla.edu}%
\thanks{$^{2}$Amazon Alexa AI. Email: \{gxiaofen, qzgao, ssshakia, thattg, sukhatme\}@amazon.com}
\thanks{$^{3}$Department of Computer Science, USC Viterbi School of Engineering.  Email: gaurav@usc.edu}
\thanks{Digital Object Identifier (DOI): see top of this page.}
}

\newcommand*{\addFileDependency}[1]{
  \typeout{(#1)}
  \@addtofilelist{#1}
  \IfFileExists{#1}{}{\typeout{No file #1.}}
}
\makeatother

\begin{document}

\maketitle
\markboth{IEEE Robotics and Automation Letters. Preprint Version. Accepted August, 2023}
{Gong \MakeLowercase{\textit{et al.}}: LEMMA: Learning Language-Conditioned Multi-Robot Manipulation} 

\begin{abstract}

Complex manipulation tasks often require robots with complementary capabilities to collaborate. We introduce a benchmark for \underline{L}anguag\underline{E}-Conditioned \underline{M}ulti-robot \underline{MA}nipulation (\benchmark) focused on task allocation and long-horizon object manipulation based on human language instructions in a tabletop setting. \benchmark features 8 types of procedurally generated tasks with varying degree of complexity, some of which require the robots to use tools and pass tools to each other. For each task, we provide 800 expert demonstrations and human instructions for training and evaluations. \benchmark poses greater challenges compared to existing benchmarks, as it requires the system to identify each manipulator's limitations and assign sub-tasks accordingly while also handling strong temporal dependencies in each task. To address these challenges, we propose a modular hierarchical planning approach as a baseline. Our results highlight the potential of \benchmark for developing future language-conditioned multi-robot systems.

\end{abstract}

\begin{figure*}[h]
    
     \centering
     \begin{subfigure}[b]{\textwidth}
         \centering
         \includegraphics[width=\textwidth]{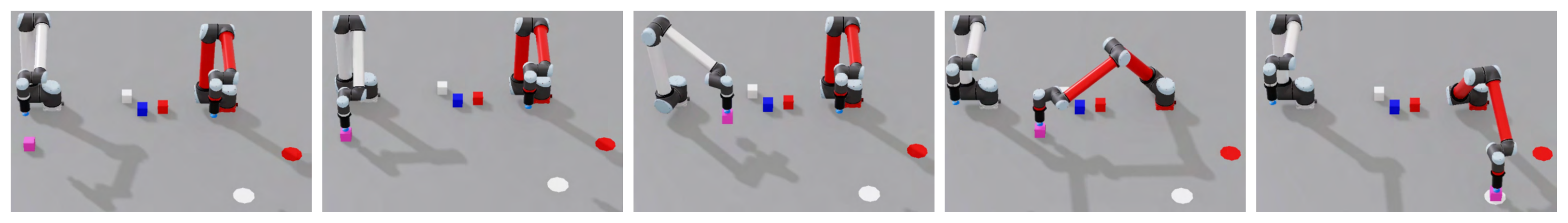}
         \caption{Pass. Instruction: Place the pink cube on the white pad. Robots: UR10 and UR10.}
     \end{subfigure}
     \begin{subfigure}[b]{\textwidth}
         \centering
         \includegraphics[width=\textwidth]{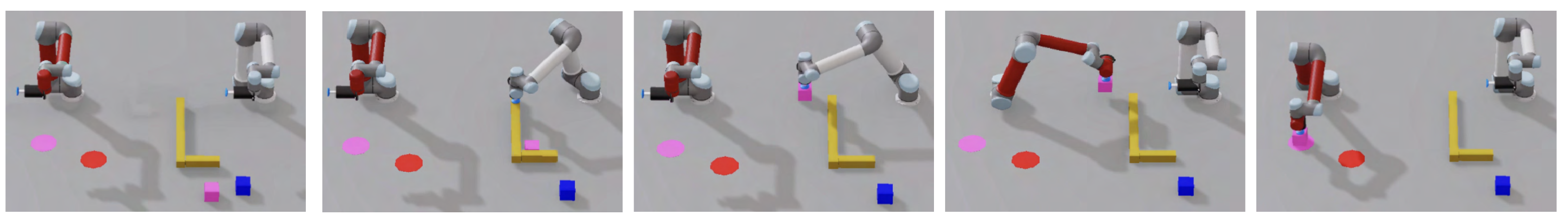}
         \caption{Hook. Instruction: Place the pink cube on the pink pad. Robots: UR5 and UR5.}
     \end{subfigure}
     \begin{subfigure}[b]{\textwidth}
         \centering
         \includegraphics[width=\textwidth]{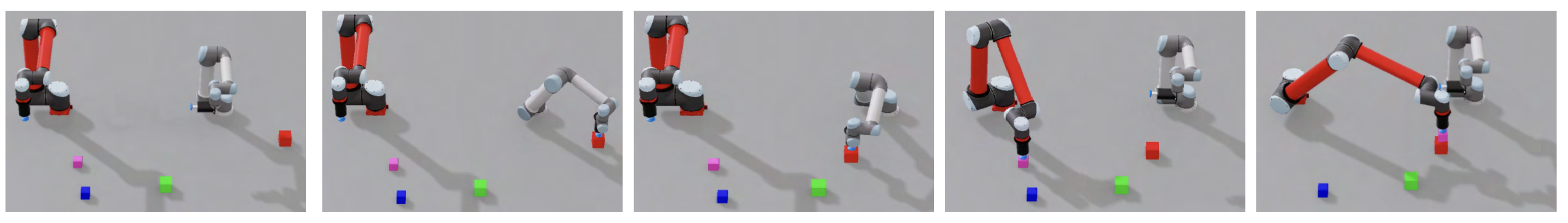}
         \caption{Stack. Instruction: Place the pink cube on top of the red cube. Robots: UR5 and UR10.}
         \label{fig:explain_c}
     \end{subfigure}
     \begin{subfigure}[b]{\textwidth}
         \centering
         \includegraphics[width=\textwidth]{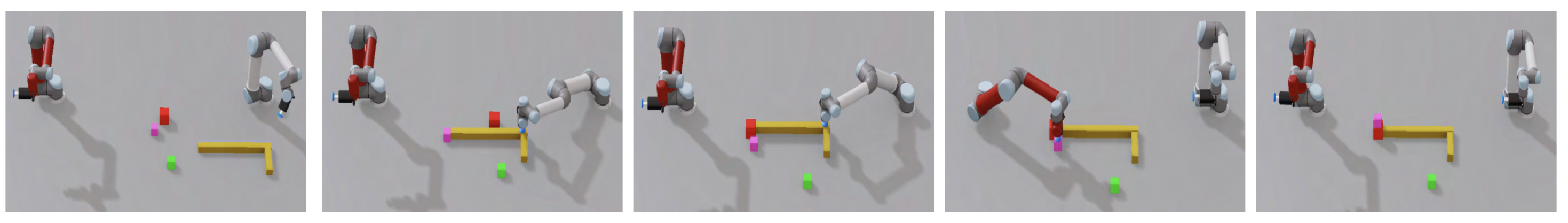}
         \caption{Poke2stack. Instruction: Place the pink cube on top of the red cube. Robots: UR5 and UR5.}
     \end{subfigure}
    \caption{Expert demonstrations and high-level instructions of tasks in \benchmark. We use minimal instructions that specify the goal, thus, it is possible that two different tasks may have the same instruction. E.g., tasks in (c) and (d) have the same instruction, but (d) requires robots to make use of the tool to poke the blocks so that the red robot can reach them. Note that the pair of robots involved in each demonstration can be different. The pair of robots in homogeneous settings (e.g., a, b, and d) have the same reach, while in heterogenous cases the reach can be different for each robot (e.g., c).}
    \vspace{-10pt}
    \label{fig:explain}
\end{figure*}


\section{Introduction}\label{sec:intro}

\IEEEPARstart{T}{here} is growing interest in connecting human language to robot actions, particularly in single-agent systems \cite{shridhar2022cliport, mees2021calvin, zheng2022vlmbench, anderson2018vision, shridhar2023perceiver}. However, there remains a research gap in enabling multi-robot systems to work together in response to language input.


Recent vision and language tasks have primarily focused on navigation and object interactions \cite{anderson2018vision, shridhar2020alfred, gao2022dialfred}. However, the lack of physical manipulation in these works makes the settings oversimplified. Although some recent studies, such as \cite{shridhar2022cliport, shridhar2023perceiver}, address vision and language object manipulation in single-robot settings, the language instructions provided specify only short-term goals, neglecting long-term objectives. \cite{huang2022inner} attempt to address these limitations by exploring long-horizon planning with manipulation for individual robots. Nevertheless, there remains a need to investigate multi-robot systems capable of accomplishing a broader range of long-horizon tasks while following language instructions.

Learning policies for multi-robot systems introduces distinct challenges, including diverse capabilities arising from physical constraints such as the location and reach of different robots. Moreover, task planning heavily depends on the spatial and physical relations between the objects and robots, in addition to the geometries of the objects. To ensure suitable task assignments, an awareness of each robot's specific physical capabilities is needed.

To tackle the language-conditioned vision-based multi-robot object manipulation problem, we have developed \benchmark, a benchmark that contains 8 types of collaborative object manipulation tasks with varying degrees of complexity. Some tasks require the robot to use tools for object-object interactions. For each task, the object poses, appearances, and robot types are randomized, requiring object affordance estimation and robot capability understanding. To enable multi-task learning, each task is paired with an expert demonstration and several language instructions specifying the task at different granularities. As a result, \benchmark introduces a diverse range of challenges in multi-robot collaboration, including physics-based object manipulation, long-horizon task planning, scheduling and allocation, robot capability and object affordance estimation, tool use, and language grounding. Each aspect poses distinct challenges and is crucial for a multi-robot system that follows human instructions to complete tasks. To evaluate existing techniques on \benchmark, we further provide several baseline methods and compare their performance to each other. We assess task performance by utilizing the latest language-conditioned policy learning models. Our results indicate that current models for language-conditioned manipulation and task planning face significant challenges in \benchmark, especially when dealing with complex human instructions. 

We make the following contributions:
\begin{itemize}[leftmargin=*]
\item We design eight novel collaborative object manipulation tasks involving robots with different physical configurations implemented in Nvidia Omniverse - Isaac Sim.
\item We provide an open-source dataset comprising 6,400 expert demonstrations and natural language instructions, including human and high-level instructions.
\item We implement a modular hierarchical planning approach as a baseline, which integrates language understanding, task planning, task allocation, and object manipulation. 
\end{itemize}


\section{Related Work}
\noindent \textbf{Language Conditioned Manipulation.}
Recent research has shown a growing interest in connecting human language to robot actions \cite{mees2021calvin, lynch2020language, shridhar2020alfred, zheng2022vlmbench, shridhar2023perceiver, zeng2022socratic, stepputtis2020language, nair2022learning}. However, these studies typically use a single-robot setting. In contrast, our work emphasizes multi-robot collaboration. In particular, our task settings are designed to require the collaboration of two robots to successfully complete the task. In contrast, in \cite{tan2020multi, liu2022embodied}, a single agent can still finish the task although multiple agents are available. In our setting, due to the limited workspace reachability of each robot, each target object can only be manipulated by one robot initially, making collaboration necessary to achieve the task goals.

\noindent \textbf{Visual Multi-Agent Collaboration.}
Visual multi-agent collaboration has attracted attention in recent embodied AI research \cite{tan2020multi, jain2020cordial, jain2019two, wang2021collaborative, chen2019visual, liu2022multi, liu2022embodied}. However, among the works that involve object manipulation, simplified non-physics based atomic actions are often employed as an abstraction for manipulation. These works often use a magic glove to attach an object to the gripper as long as hand-crafted conditions are met. For example, an object within 15cm to the gripper can be automatically snapped to the gripper \cite{szot2021habitat}. While this simplification does not affect learning robot task planning, it is unsuitable for learning low-level manipulation policies. We do not make such a simplification here instead using a gripper to interact with objects physically. Additionally, most previous works study the collaboration problem in a single-task setting. In contrast, our work uses a multi-task setting requiring the comprehension of a textual description to understand the goal.

\begin{table*}[t!]
\centering
\small
\resizebox{\textwidth}{!}{%
\begin{tabular}{ccccccccccc}
\toprule
Benchmark                   & Alfred\cite{shridhar2020alfred}  & MQA\cite{deng2020mqa} & Calvin\cite{mees2021calvin} & M-EQA\cite{tan2020multi} & Ravens\cite{zeng2021transporter} & Vlmbench\cite{zheng2022vlmbench} & CH-MARL\cite{sharma2022ch}& TBP \cite{jain2019two}& EMATP\cite{liu2022embodied} &\textbf{\benchmark} \\
\midrule
Language                    & \cmark  & \cmark & \cmark & \cmark & \xmark & \cmark & \cmark  & \xmark& \cmark&\cmark \\
Multi-task & \cmark & \cmark & \cmark & \cmark &\xmark & \xmark & \xmark &\xmark & \cmark &\cmark \\
Manipulation & \xmark & \cmark & \cmark & \xmark & \cmark & \cmark & \xmark  &\xmark &\xmark&\cmark \\
Multi-agent & \xmark & \xmark & \xmark & \cmark & \xmark & \xmark & \cmark &\cmark & \cmark&\cmark \\
Tool Use & \cmark & \xmark & \xmark & \xmark & \xmark & \xmark & \xmark &\xmark & \cmark&\cmark \\
Temporal Dep. & \cmark & \xmark & \xmark & \xmark & \xmark & \xmark & \xmark & \xmark & \cmark&\cmark \\

\bottomrule
\end{tabular}
}
\caption{\textbf{Comparison with other benchmarks.} \benchmark evaluates the performance of language-conditioned multi-agent object manipulation in long-horizon tasks. \textbf{Multi-task}: using a multi-task setting. \textbf{Language}: language instructions to specify goal. \textbf{Manipulation}: physical object manipulation. \textbf{Multi-agent}: requiring multiple agents for task completion. \textbf{Tool use}: requiring the robot to use a tool to interact with other objects. \textbf{Temporal Dep}: temporal dependency between sub-tasks. }
\vspace{-10pt}
\label{tab:comparison}
\end{table*}

\noindent \textbf{Bimanual Robot Manipulation.} There is a rich set of literature on bimanual robot manipulation \cite{chen2022towards, takata2022efficient, stavridis2021pick, smith2012dual, zhang2019robot, stepputtis2022system, lertkultanon2018certified}. These works address important problems in dual-arm coordination with a focus on coordinated control and collision avoidance. However, there is less exploration of multi-robot task planning and allocation for long-horizon tasks with strong temporal dependencies, along with workspace management. In addition, these works typically do not involve vision and language inputs, especially for the recognition of the physical limitations of different robots from vision input. More importantly, previous research usually employs robot arms of the same type. In contrast, our work considers the settings of both heterogeneous and homogeneous robot arms. 

\noindent \textbf{Visual Robot Task and Motion Planning.}
Traditionally, most works in this area use search over pre-defined domains for planning, which require extensive domain knowledge and accurate perception. Moreover, they often scale poorly with an increasing number of objects. Another line of work involves generating task and motion plans given scene images \cite{driess2020deep, 21-driess-IJRR}. In contrast, in our settings, the model uses both RGBD images and textual descriptions as input for multi-task learning. Most recently, \cite{singh2022progprompt} generated robot policies in the form of code tokens using large language models (LLMs). Nonetheless, they only focus on single-agent task planning. We compare and contrast our benchmark with existing works in \Cref{tab:comparison}.



\section{Problem Formulation}
\label{sec: formulation}
Assume a robot system  comprised of $N$ robots that is tasked to complete a complex manipulation task, the goal of which is specified by a language instruction $x_L=\{x_{l}\}_{l=1}^{L}$, which is a sequence of $L$ word tokens. The full task can be decomposed into $M$ sub-tasks $d_M = \{d_{m}\}_{m=1}^{M}$. $d_M$ represents the full task, and $d_m$ represent each sub-task in $d_M$. Given the language instruction $x_L$, our goal is to find a valid and optimal sub-task allocation. We define $q_{im}$ and $c_{im}$ as the quality and cost, respectively, for allocating robot $i$ to work on sub-task $m$. Then the combined utility for the sub-task is:
$$
u_{im}= \begin{cases}q_{im}-c_{im}, & \text{ if robot } i \text{ can execute sub-task m} \\ -\infty. & \text { otherwise }\end{cases}
$$
We define the assignment of sub-task $m$ to robot $i$ as 
$$
v_{im}= \begin{cases}1, & \text{robot } i \text{ is assigned to sub-task m} \\ 0. & \text { otherwise }\end{cases}
$$
with $\gamma_m = i$ being an assignment variable for each sub-task $m$, indicating that sub-task $m$ is assigned to robot $i$. 

The goal is to maximize the utility of the full manipulation task under a time constraint. Defining the execution time for task $m$ by robot $i$ as $\tau_{i m}$, and the maximum time allowed to execute the task as $T_{max}$, we can express the task decomposition and assignment problem as follows:

\begin{equation}
\label{equ:obj}
\argmax_{v} \sum_{i=1}^{N} \sum_{m=1}^{M} u_{i m} v_{i m}
\end{equation}

Subject to:
$$
\begin{array}{rlrl}
\sum_i\sum_m  \tau_{i m} v_{i m} & \leq T_{max} \\
\sum_{i} v_{i m} & \leq 1 & \forall m \in M \\
v_{i m} & \in\{0,1\} & \forall i \in N, \forall m \in M
\end{array}
$$

As pointed out by \cite{korsah2013comprehensive}, this problem cannot be solved in polynomial time. In this work, we tackle this problem by learning from expert demonstrations, so that each sub-task can be assigned to a capable robot to ensure successful task execution. With the sub-task $d_m$ and its assignment $\gamma_m$, an object manipulation policy $\pi(s^{t+1} | s^{t}, o_N^t, x_L, d_{m}, \gamma_{m})$ can be used to move a specific object from its current pose $s^{t}$ to its target pose $s^{t+1}$, given the observations $o_N^t = \{o^{t}_{i}\}_{i=1}^{N}$ and language instruction $x_L$. In this work, the observation $O_N$ consists of robot joint configurations, RGBD images associated with each robot, and camera parameters.

\begin{table*}[h!]
\centering
\small
\resizebox{\textwidth}{!}{%
\begin{tabular}{lcccccccc}
    \hline
    Task Type & Pass  & Pass2 & Stack & Stack2 & Poke & Poke\&Stack & Hook & Hook\&Stack \\ \hline

    Objects Categories & Cube, Pad & Cube, Pad  & Cube & Cube &\makecell{Cube, Pad, \\ Stick}  & Cube, Stick &  \makecell{Cube, Pad, \\ Stick} & Cube, Stick \\
    Using Tools & \xmark & \xmark &\xmark  & \xmark & \cmark & \cmark & \cmark  & \cmark \\
    Passing Tools & \xmark & \xmark &\xmark  & \xmark & \xmark &\xmark  & \xmark & \cmark \\
    Number of Objects & 3-6 & 6-8 & 2-5 & 4-6 & 4-6 & 3-5 & 3-5 & 3-6 \\
    Number of Sub-tasks & 2 & 4 & 2 & 4 & 3 & 5 & 4 & 7 \\ \hline

\end{tabular}%
}

\caption{There are 8 types of tasks in \benchmark. Tasks require 2-7 sub-tasks, with each sub-task requiring the robot to pick up an object, move to a location and put it down. Some tasks require the robots to use tools. In addition, the most difficult task, Hook\&Stack, requires passing the tool from one robot to the other.}
\vspace{-10pt}
\label{tab:task_feature}
\end{table*}

\section{\benchmark Benchmark}

We introduce \benchmark to address the language-conditioned multi-robot manipulation problem. This benchmark is designed to evaluate a system's ability to dynamically perform task allocation and object manipulation in a tabletop environment. \benchmark sets itself apart from existing language-conditioned robotic manipulation benchmarks, such as \cite{zeng2021transporter, zheng2022vlmbench, mees2021calvin}, in several key aspects:

\begin{itemize}[leftmargin=*]
\item All tasks in \benchmark feature strong temporal dependencies, making the execution order of sub-tasks critically important. Out-of-order execution will result in task failure.
\item \benchmark tasks are exclusively multi-agent based. Due to the robots' reachable space limitations, it is impossible to complete tasks in \benchmark using a single agent.
\item As illustrated in \Cref{fig:explain}, robots are provided with only minimal high-level instructions. This requires the model to have a deep understanding of the environment and plan a sequence of actions accordingly to reach the goal specified by the instruction. Using a language classifier to determine the task type and a template to form task plans, as in \cite{min2021film}, is inadequate, as multiple tasks may share the same language instructions.
\item \benchmark allows robots with different physical configurations to collaborate on manipulation tasks. Two types of robots are provided in \benchmark, namely UR10 and UR5.
\end{itemize}
A detailed comparison between \benchmark and other related benchmarks is shown in \Cref{tab:comparison}.

\subsection{Task settings}
In \benchmark, we focus on tasks that require a multi-robot system to complete, given a single instruction and visual observations from top-down cameras placed above each robot. Specifically, we consider a two-robot setting under centralized control. The high-level goal is specified by a single natural language instruction such as "Place the red block on top of the white pad". However, this instruction may not provide all the necessary details on how to complete the task. To thoroughly assess system performance in language-conditioned multi-robot collaboration, we have designed 8 tasks of varying difficulty. The tasks are implemented in a simulated tabletop environment in NVIDIA Omniverse Isaac-Sim. Task statistics can be found in \Cref{tab:task_feature}.

\noindent{\textbf{Pass}.} The first robot is required to pick up a cube of a designated color in its own reachable workspace and place it within a shared workspace, i.e. a space that is reachable by both robots. Following this, the second robot must pick up the designated cube and position it on top of a specified pad in its reachable workspace.

\noindent \textbf{Stack}. The first robot picks up a designated cube in its reachable workspace and places it in the shared workspace. Subsequently, the second robot picks up a different designated cube and stacks it on top of the first cube.

\noindent \textbf{Poke}. Initially, the cube is unreachable by either robot. One robot has access to a tool and must use it to poke the designated cube into the other robot's reachable workspace. The second robot then picks up the cube and places it in the target pad in its reachable workspace.

\noindent \textbf{Hook}. Initially, the cube is unreachable by either robot. However, one of the robots can use a tool to hook the cube into its own reachable space. After that, the robot need to move it to the shared workspace so that the other robot can pick it up and position it on the target pad.

\noindent \textbf{Pass2}. As an extension of Pass, the task requires the robots to pass two different objects to each other and place them in their respective target locations. 

\noindent \textbf{Stack2}. As an extension of Stack, the task requires the robots to construct two block towers, each requires two cubes of different designated colors. 

\noindent \textbf{Poke\&Stack}. The task requires one robot to use a tool to poke two designated cubes into the other robot's workspace. The other robot then places one cube on top of the other.

\noindent \textbf{Hook\&Stack}. The task requires one robot to use a tool to hook a designated cube into its own reachable space and pass the tool to the second robot, allowing it to perform its hook action on the second cube. The first robot then moves the first hooked cube to the shared workspace so that the second robot can pick it up and stack it on the second block.



\begin{table*}[t]
    \centering
    \begin{tabular}{|l|c|c|}
    \hline
    Task Type & High-Level Instruction & Human Instruction \\
    \hline
    Pass & place the \textit{pick-color} cube on top of the \textit{place-color} pad & \makecell{pick a \textit{pick-color} cube from one side, \\put it at the center and put it on the \textit{place-color} circle} \\ 
    \hline
    Pass2 & \makecell{place the \textit{pick-color1} cube on top of the \textit{place-color1} pad\\ and place the \textit{pick-color2} cube on top of the \textit{place-color2} pad} & \makecell{ place the \textit{pick-color1} cube on the \textit{place-color1} \\ pad and \textit{pick-color2} cube on the \textit{place-color2} pad \\ in the opposite direction }\\
    \hline
    Stack & place the \textit{pick-color} cube on top of the \textit{place-color} cube & \makecell{take \textit{place-color} block and place it in center \\ under the \textit{pick-color} block} \\
    \hline
    Stack2 & \makecell{place the \textit{pick-color1} cube on top of the \textit{place-color1} cube\\ and place the \textit{pick-color2} cube on top of the \textit{place-color2} cube} &\makecell{assemble two block towers by placing\\ the \textit{place-color1} cube under the \textit{pick-color1} cube \\ and the \textit{place-color2} cube under the \textit{pick-color2} cube} \\
    \hline
    Poke & place the \textit{pick-color} cube on top of the \textit{place-color} pad & \makecell{use the L object to push the \textit{pick-color} block \\ to the other robot and place it on the \textit{place-color} pad} \\
    \hline
    Poke\&Stack & place the \textit{pick-color} cube on top of the \textit{place-color} cube & \makecell{grab the brown L to push the \textit{pick-color} and \\ \textit{place-color} cubes closer to the other robot, \\ so it can grab and place the \textit{pick-color} cube \\ on top of the \textit{place-color} cube}\\
    \hline
    Hook & place the \textit{pick-color} cube on top of the \textit{place-color} pad & \makecell{fetch the \textit{pick-color} cube using the L shaped object \\ and place it on top of the \textit{place-color} pad}\\
    \hline
    Hook\&Stack & place the \textit{pick-color} cube on top of the \textit{place-color} cube & \makecell{use the grippers to pick up the \textit{pick-color} block \\ from the hook and stack it on the \textit{place-color} block} \\
    \hline
    
    \end{tabular}
    \caption{Example high-level instruction and 
crowd-sourced human language instruction templates to specify manipulation tasks in \benchmark. \textit{pick-color} and \textit{place-color} represent the color of objects being picked up and placed on.}
\vspace{-15pt}
\label{tab:instruction}
\end{table*}

\subsection{Action Space and Observation Space}

The default action space of each robot is the end-effector position. Nevertheless, alternative control mechanisms such as joint positions, joint velocities, and joint torques can also be accommodated for controlling 6 degrees-of-freedom robots. Due to physical constraints, each robot has a unique reachable workspace, allowing it to only interact with a limited set of objects in the task. 

To obtain visual observations, we place a fixed RGBD camera above each robot. Each camera has a limited field of view and cannot capture all the objects. We follow \cite{shridhar2022cliport} to process the vision input: we first generate the scene point cloud based on all RGBD images and camera parameters, and use the point cloud to obtain the top-down orthographic RGBD reconstruction of the scene. 

\vspace{-8pt}

\subsection{Task Generation}

 For diversity among the tasks, we use a rejection sampling mechanism to generate task instances. Each task instance specifies the initial environment configurations and goal conditions. To ensure that the task completion requires both robots, we make sure that the initial location of the target object falls into the reachable workspace of only one robot. In addition to the target object, we also add some distractor objects to make the task more challenging: 
\begin{enumerate}[leftmargin=*]
    \item Specify each robot's type, location, and color. There are two robot types, UR5 and UR10, respectively. Each robot has two different colors, red and white. Colors and types are randomly assigned to each robot.  
    \item Sample the goal condition of the task, including the colors of the target objects and their goal locations, according to the task type. E.g., the goal condition for the \textit{Stack pink on red} task is satisfied when a pink block is on top of a red block, as shown in \Cref{fig:explain_c}. There are five available colors for blocks and pads, including pink, red, white, blue, and green. The color of the tool is always yellow. 
    \item Sample the initial locations of target objects, so that each object is only within the reachable space of a single robot.
    \item Sample the colors and locations of distractor objects while making sure the colors of the target objects are unique. 
\end{enumerate}
The above sampling mechanism is repeated until one valid task instance is generated. 

\subsection{Expert Demonstration}

Given a task specification, we use an oracle task and motion planner to create expert demonstrations. Based on the initial configurations and goal conditions of the task, the oracle creates a task plan consisting of a sequence of sub-tasks $d_M$, and the allocated robot $\gamma_m$ for each sub-task $d_m$. The sub-task follows a pick-and-place procedure, specifying the target object to pick up and the target pose at which to place the object. Once the task plan is generated, an RMPflow motion planner is utilized to generate a motion plan based on ground truth object locations at each time step. All the motion plans are executed by the designated robot, and the corresponding RGBD images and camera poses are recorded to form the expert demonstration data. 


\subsection{Language Instructions}

 We assign high-level instruction and a human instruction to each task instance.  For high-level instruction, we manually define a template for each task and lexicalize the template using the goal of the task. For human instruction, we first crowd-sourced 500 templates on Amazon Mechanical Turk and selected 80 valid templates that are not too verbose and uniquely specify the goal given the visual observation, 10 for each task type. For each task instance, we then sample a template from the template pool and lexicalize it to generate human instruction. Some examples of the instruction templates can be found in \Cref{tab:instruction}. 

In summary, each datapoint in \benchmark contains the following information: 1) a manipulation task, specified by the initial configurations and goal conditions, 2) an expert demonstration, including camera poses and RGBD images at each timestep, 3) a high-level instruction and a human instruction specifying the task. As a result, we generate 800 data sessions for each task type, with a total of 6400 sessions. For each task type, we keep 700 sessions in the \textit{training} set, 40 in the \textit{validation} set, and 60 in the \textit{test} set.

\subsection{Evaluation Metrics}
We have adopted success rate as the evaluation metric in \benchmark. Task success is defined as 1 if the task goal-conditions are met at the end of the episode, and 0 otherwise. The time constraint of each episode $T_{max}$ is set to 100 seconds. The task specified by the instruction \textit{Place the red block on the green circle}, for example, is considered successful if, at the end of the episode, the red block is on top of the green circular pad. 

\section{Baseline Models}

\begin{figure*}[h]
    \centering
    \begin{subfigure}{0.5\textwidth}
        \centering
        \includegraphics[width=\textwidth]{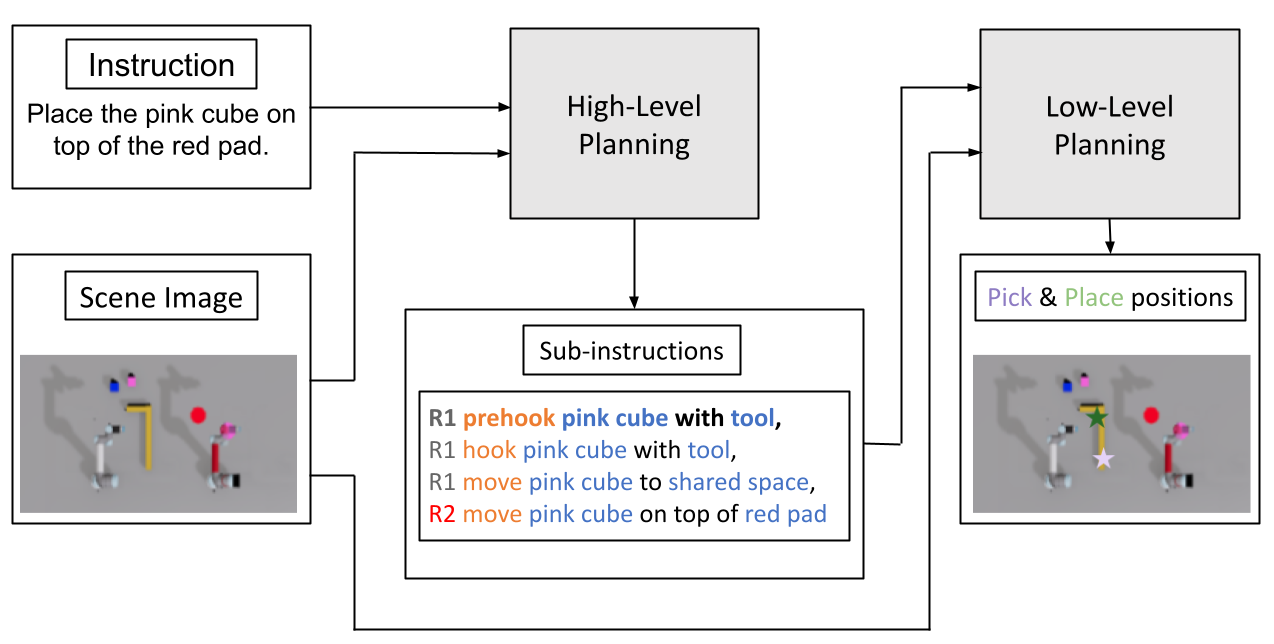}
        \caption{Overall architecture of the baseline model.}
        \label{fig:architecture}
    \end{subfigure}
    \quad
    \begin{subfigure}{0.4\textwidth}
        \centering
        \includegraphics[width=\textwidth]{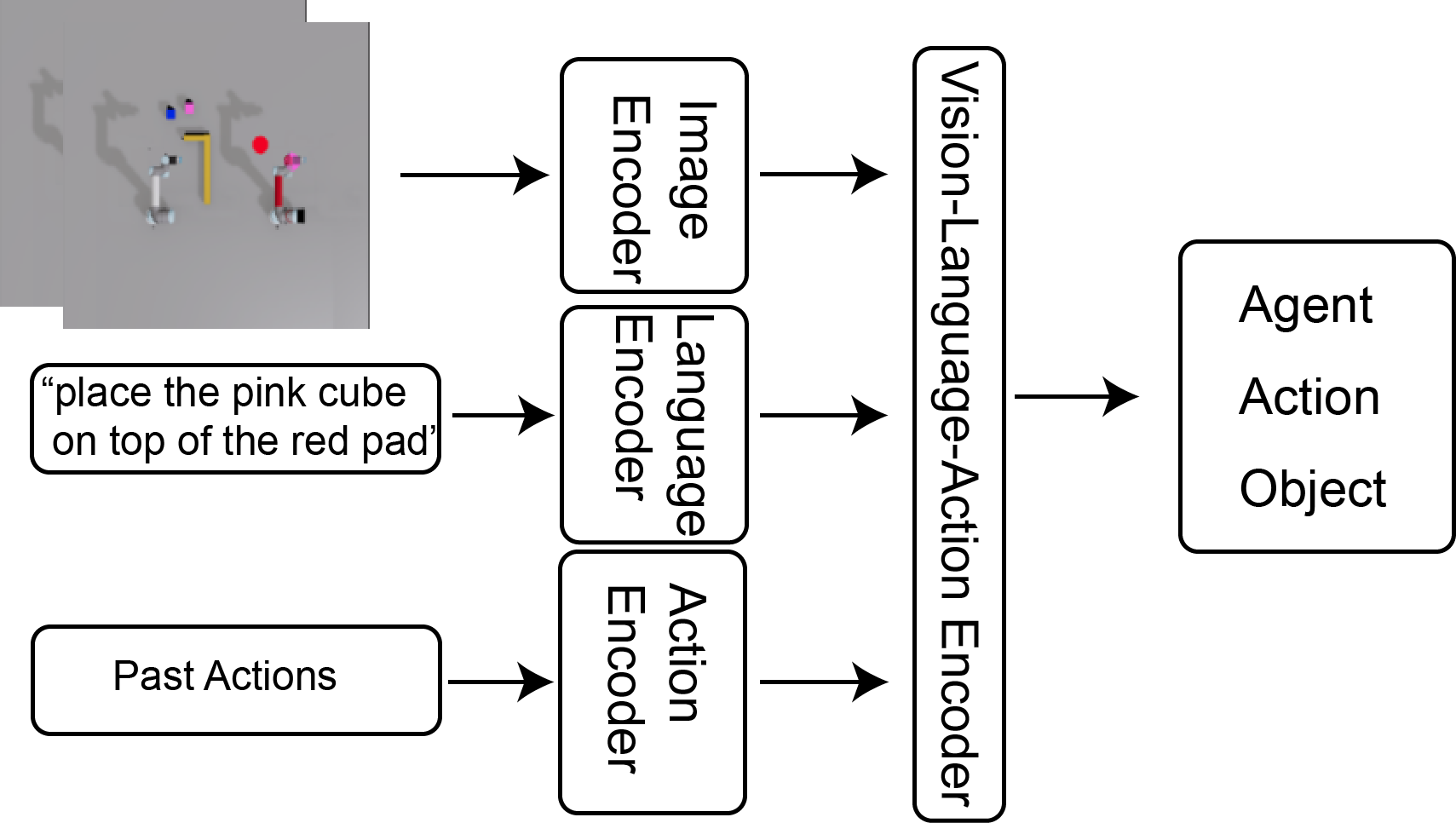}
        \caption{The architecture of the Episodic Transformer model used as the high-level planner.}
        \label{fig:et_architecture}
    \end{subfigure}
    \caption{ Our baseline model involves a high-level task planning module and a low-level planning module. The high-level planning module takes as input the top-down fused scene image and human instruction to generate sub-instructions that assign the corresponding sub-tasks to robots based on their limitations. Each sub-task is specified by action primitives (in orange) and objects (in blue). The low-level planning module uses the sub-instruction and top-down projection of the scene to generate pick and place locations (visualized as purple and green stars respectively) of the gripper in the scene.}
    
    \vspace{-10pt}
\end{figure*}

Consider the problem of learning multi-agent task and motion planning with two robots given a single language instruction and visual observations. The problem can be decomposed into two sub-problems: (a) multi-robot task planning and task allocation, and (b) single-robot planning given the assigned sub-task. To this end, we design a modular baseline model which includes a high-level planner for deciding which sub-task a robot should work on and a low-level planner for generating pick and place locations for the gripper given the assigned sub-task. \Cref{fig:architecture} shows the overall architecture of our baseline model. We follow \cite{shridhar2022cliport} to use the fused point clouds generated from the RGB and depth images of two cameras. Then we use the top orthographic projection of the point cloud as the visual input. 

\subsection{Action Primitives}


We define six action primitives: \textit{move, prehook, hook, prepoke, poke} and \textit{stop}. The non-stop action primitives follow generalized pick-and-place settings. \textit{Move} is the standard pick-and-place primitive, requiring the robot to pick up the object and move it to a specified location. \textit{Prehook} and \textit{prepoke} require the robot to pick up the tool and align it with the target object to prepare for \textit{hook} and \textit{poke} respectively. The required height of the gripper is different for each primitive. For \textit{hook} and \textit{poke}, the height of the gripper after picking is set at 5cm, enabling the tool to come into contact with the target object. For other primitives, the height of the gripper is set to 30cm to avoid contact with other objects between the pick and place actions. The sequence of action primitives required to complete the task depends on the relative poses of target objects, such as cubes, tools, and pads. As a result, the same language instruction can correspond to completely different sequences of primitive actions, depending on the specific arrangement of these objects.

\subsection{Multi-Robot Task Planning and Task Allocations}

The multi-robot collaboration task can be decomposed into a sequence of sub-tasks, each can be completed by a single robot. The sub-task allocation can be represented by a tuple $(d_m^t, g_m^t, \gamma_m^t)$. $g_m^t=(e, p, q)$ represents the entities to specify the sub-task $d_m^t$, including the action primitive $e$, the pick entity $p$ and place entity $q$. E.g. $(\textit{\text{Move}}, \textit{\text{Red Cube}}, \textit{\text{Shared Space})}$ indicates moving a red cube on top of the shared workspace between two robots. $\gamma_m^t$ is the sub-task assignment indicating which robot is to perform the task. $(d_m^t, g_m^t, \gamma_m^t)$ are further lexicalized using templates to form the sub-instruction specifying the sub-task and its allocation (\Cref{fig:architecture}). In practice, we use Episodic Transformer \cite{pashevich2021episodic}, a vision and language task planner to generate the sub-instructions. The approach is a language-conditioned visual task planning method that employs a transformer architecture for long-horizon planning. As shown in \Cref{fig:et_architecture}, ET uses the historical visual and language information in the entire episode to capture long-term dependencies between actions. It leverages the transformer architecture to first separately encodes image histories, language instructions, and past action histories, and then perform cross-attention across modalities to decode robot assignment, action primitives, and the target object separately.

The choice between neural-based open-loop planning and closed-loop planning is often debatable. Open-loop planning cannot adapt to errors made in the planning process. However, it is more stable to train since the training distribution is often more aligned with the testing distribution. Here we consider both open-loop planning and close-loop planning for our benchmark and compare their performance. Note the original Episodic Transformer is closed-loop only and requires new observation at each time step to plan the next action (i.e. \textit{Single-step}). We modify the algorithm to use only the initial visual observation by providing it as input repeatedly during the loop to plan the whole sub-instruction sequences (i.e. \textit{Multi-step}).

\vspace{-10pt}

\begin{figure}
    \centering
    \includegraphics[width=0.75\linewidth]{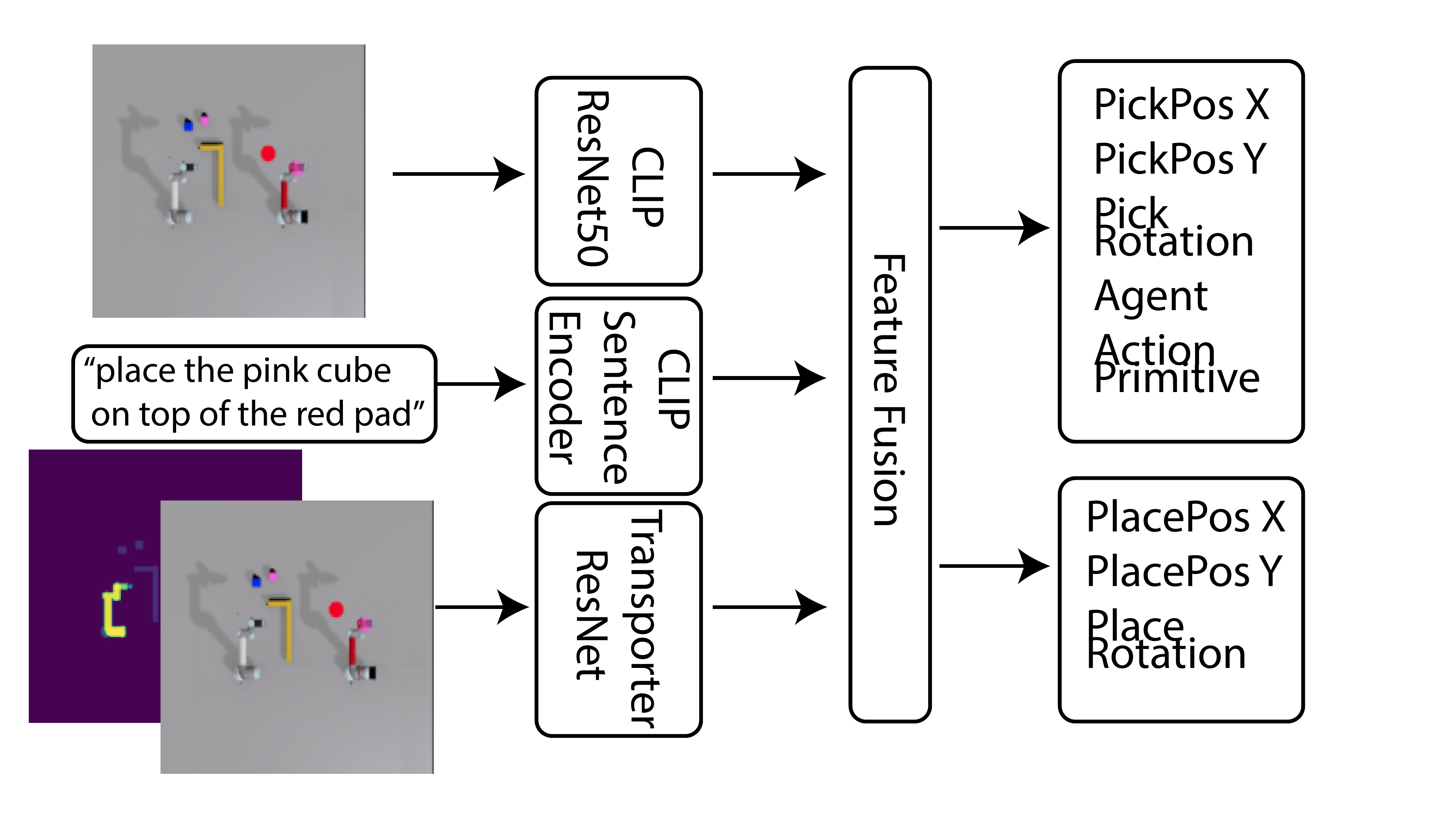}
    \caption{Multi-Agent Cliport Model architecture. The output is extended to include robot assignment and action primitive for each predicted action.}
    \label{fig:cliport}
\end{figure}

\subsection{Single Agent Object Manipulation and Grounding}

In this work, we use CLIPort \cite{shridhar2022cliport} as the low-level planner. Formally, at each time step, the algorithm focuses on learning a goal-conditioned policy $\pi$ that produces actions $\mathbf{a}^t$ based on the current visual observation ${o}^t$ and a language instruction $x_{L}^{t} = \{x^{t}_{l}\}_{l=1}^{L}$. The visual observation ${o}^t$ is an orthographic top-down RGB-D reconstruction of the scene, where each pixel corresponds to a point in 3D space. As shown in \Cref{fig:architecture}, in our use case, each input language instruction $x_{L}^{t} = (g_m^t, \gamma_m^t)$ specifies a sub-task being allocated to robot $\gamma_m^t$ at time step $t$. As a result, the goal-conditioned policy is defined as 
$$
\pi\left( o_N^t, x_{L}^{t} \right) = \pi\left( o_N^t, g_m^t, \gamma_m^t \right) \rightarrow \mathbf{a}^t=\left(\mathcal{T}_{\text {pick }}, \mathcal{T}_{\text {place }}\right) \in A, 
$$ 
where the actions $\mathbf{a} = (\mathcal{T}{\text{pick}}, \mathcal{T}{\text{place}})$ denote the end-effector poses for picking and placing respectively. CLIPort is designed for tabletop tasks, with $\mathcal{T}{\text{pick}}, \mathcal{T}{\text{place}} \in \mathbf{SE}(2)$.

\vspace{-5pt}

\begin{table*}[h!]
\centering
\begin{tabular}{lccccccccc}
    \hline
    Task Type & Pass & Pass2 & Stack & Stack2 & Poke & Poke\&Stack & Hook & Hook\&Stack & Avg \\
    \hline
    M-CLIPort & 34.33$\pm$2.19 & 0.00$\pm$0.00 & 10.67$\pm$0.82 & 0.00$\pm$0.00 & 15.33$\pm$4.64 & 6.00$\pm$0.82 & 21.67$\pm$4.34 & 1.67$\pm$1.50 & 11.21$\pm$0.76\\
    Single-step & 87.00$\pm$0.67 & 27.00$\pm$1.25 & 41.67$\pm$0.00 & 30.00$\pm$3.86 & 28.00$\pm$3.86 & 10.33$\pm$1.24 & 86.70$\pm$2.36 & 35.00$\pm$2.36 & 43.21$\pm$0.45\\
    + GTA & \textbf{100.00$\pm$0.00} & \textbf{84.67$\pm$2.67} & \textbf{91.33$\pm$0.67} & 72.00$\pm$0.67 & 46.67$\pm$0.00 & 32.33$\pm$1.70 & \textbf{98.33$\pm$0.11} & \textbf{73.33$\pm$2.36} & \textbf{74.83$\pm$0.44}\\
    Multi-step & 83.06$\pm$1.15 & 28.06$\pm$1.78 & 28.89$\pm$0.79 & 31.11$\pm$2.48 & 30.83$\pm$2.31 & 9.17$\pm$1.27 & 82.78$\pm$1.57 & 24.72$\pm$4.66 & 39.83$\pm$0.84\\
    + GTA & \textbf{100.00$\pm$0.00} & 83.89$\pm$3.00 & 90.00$\pm$0.00 & \textbf{74.72$\pm$1.78} & \textbf{55.28$\pm$0.62} & \textbf{35.56$\pm$1.24} & 96.94$\pm$0.62 & 58.34$\pm$1.93 & 74.34$\pm$0.56\\
    \hline
\end{tabular}
\caption{Performance on the test set with \textbf{high-level} instructions. M-CLIPort: Multi-agent Cliport. Single-step: high-level planning generates each sub-instruction based on the new observation. Multi-step: high-level planning generates the complete sub-instruction sequences from the initial observation. GTA:  replacing the robot task allocation results from either single-step or multi-step planning by the ground truth while preserving the predicted action primitives and objects.}
\vspace{-5pt}
\label{tab:test_result_high_new}
\end{table*}

\begin{table*}[h!]
\centering
\begin{tabular}{lccccccccc}
    \hline
    Task Type & Pass & Pass2 & Stack & Stack2 & Poke & Poke\&Stack & Hook & Hook\&Stack & Avg\\ \hline
    M-CLIPort  & 25.83$\pm$1.86 & 0.00$\pm$0.00 & 20.00$\pm$2.36 & 0.00$\pm$0.00 & 4.58$\pm$1.38 & 1.25$\pm$1.38 & 3.75$\pm$0.72 & 0.00$\pm$0.00 & 6.93$\pm$0.47 \\
    Single-step  & 44.44$\pm$2.29 & 1.67$\pm$0.00 & 21.67$\pm$1.36 & 0.00$\pm$0.00 & 13.88$\pm$1.24 & 2.50$\pm$0.83 & 25.56$\pm$1.57 & 9.17$\pm$1.60 & 14.86$\pm$0.26\\
    + GTA & 63.33$\pm$0.00 & \textbf{5.56$\pm$1.24} & 59.44$\pm$1.24 & \textbf{1.39$\pm$0.62} & \textbf{25.28$\pm$0.62} & \textbf{9.17$\pm$1.27} & \textbf{30.00$\pm$0.00}  & \textbf{19.72$\pm$0.62}  & 26.74$\pm$0.20 \\
    Multi-step  & 58.33$\pm$1.18 & 0.00$\pm$0.00 & 38.54$\pm$0.99 & 0.00$\pm$0.00 & 7.71$\pm$2.92 & 2.92$\pm$1.10 & 21.46$\pm$0.99 & 3.96$\pm$1.16 & 16.61$\pm$0.39\\
    + GTA & \textbf{81.89$\pm$0.55} & 5.00$\pm$1.44 & \textbf{70.42$\pm$0.72} & 0.00$\pm$0.00 & 15.00$\pm$0.00 & 8.33$\pm$1.67 & 23.33$\pm$0.00 & 14.17$\pm$1.87 & \textbf{27.27$\pm$0.30} \\
    \hline
\end{tabular}
\caption{Performance on the test set with \textbf{human} instructions.}
\vspace{-10pt}
\label{tab:test_result_human_seen}
\end{table*}


\begin{table}[h!]
\centering
\begin{tabular}{lccc}
    \hline
    No. Distractors & 0 & 1 & 2\\
    \hline
    Pass & \textbf{91.67$\pm$0.00} & 87.06$\pm$2.35 & 81.18$\pm$3.50 \\ 
    Pass-human & \textbf{75.00$\pm$0.00} & 53.92$\pm$2.19 & 37.25$\pm$5.55  \\ 
    Stack & \textbf{46.15$\pm$0.00} & 40.00$\pm$3.08 & 16.92$\pm$2.86 \\ 
    Stack-human & 16.67$\pm$2.87 & \textbf{28.33$\pm$2.36} & 15.38$\pm$6.28 \\ 
    Poke & \textbf{33.08$\pm$7.14} & 30.00$\pm$4.44 & 17.50$\pm$2.50 \\
    Poke-human & \textbf{16.03$\pm$2.64} & 12.96$\pm$2.62 & 11.46$\pm$6.67 \\
    Hook & \textbf{91.30$\pm$3.89} & 83.16$\pm$3.94 & 84.44$\pm$7.37 \\ 
    Hook-human & 26.09$\pm$2.51 & \textbf{32.46$\pm$3.62} & 17.59$\pm$2.07 \\ 
    \hline
\end{tabular}
\caption{Impact of distractors (single-step planning)}
\label{tab:distractors}
\end{table}

\begin{table}[h!]
\centering
\vspace{4px}
\begin{tabular}{lccc}
    \hline
    Robot Type & UR5\&UR5 & UR5\&UR10 & UR10\&UR10\\
    \hline
    Multi-step & 38.60$\pm$3.25 & 36.32$\pm$0.47 & \textbf{46.89$\pm$1.69} \\ 
    Multi-step-human & 15.71$\pm$0.59 & \textbf{19.16$\pm$0.83} & 13.06$\pm$0.53\\ 
    Single-step& 37.44$\pm$1.00 & 41.05$\pm$1.30 & \textbf{51.94$\pm$1.17} \\
    Single-step-human& \textbf{16.81$\pm$0.40} & 15.57$\pm$0.33 & 11.94$\pm$0.96 \\
    M-CLIPort & \textbf{13.50$\pm$0.84} & 11.27$\pm$0.51 & 9.10$\pm$2.23 \\
    M-CLIPort-human & 6.62$\pm$0.93 & \textbf{8.52$\pm$1.25} & 4.48$\pm$1.06 \\
    \hline
\end{tabular}
\caption{Impact of robot types}
\label{tab:robot_types}
\end{table}

\subsection{Multi-agent Cliport}
 Since the sub-instruction already contains the sub-task allocation $\gamma_m^t$ and action primitive $e$, the CLIPort module used as our low-level planner only needs to predict the pick and place location. To compare with this modular approach, we present a modified version of the original CLIPort module (i.e. M-CLIPort) to perform task planning and task allocation explicitly in an end-to-end fashion. We extend the output to a higher dimensional vector to predict the robot assignment and action primitive to use in addition to pick and place locations, as shown in \Cref{fig:cliport}.

\section{Experiments}




\subsection{Evaluation Workflow}
We train the CLIPort module for 300K steps on the training set and save a checkpoint every 20K steps. Then we perform checkpoint selection on the validation split for both the high-level planner and the low-level planner. For the low-level planner, we use the ground-truth sub-instructions as input for checkpoint selection. For the high-level planner, we train the Episodic Transformer model for 30 epochs and save a checkpoint at the end of each epoch. We choose the best checkpoint based on the accuracy of predicted task plans on the validation split. We report the performance of a combination of best-performing high-level and low-level checkpoints on the test set. Since the physics and rendering in Isaac-sim are not deterministic, we evaluate all tasks for 10 runs and report the means and standard deviations.

\vspace{-10pt}

\subsection{Experiment Results}



\subsubsection{High-level Instructions}
The results shown in \Cref{tab:test_result_high_new} compare language-conditioned policies with different task-planning modules. The M-CLIPort fails for all tasks with longer horizons. In comparison, our modular hierarchical planning approach works reasonably well for most tasks but fails for very long-horizon tasks (i.e. poke\&stack and hook\&stack). As an ablation, we further supply ground truth task allocation to the planning module (i.e. GTA). The results show that system performance can be greatly improved with ground truth task allocation. This indicates that task allocation is quite challenging since it requires understanding the reachable workspace of robots to determine which robot should be assigned to a certain sub-task. 

\subsubsection{Human Instructions}
The results shown in \Cref{tab:test_result_human_seen} demonstrate the system performance under human instructions. Human instructions are more complex than high-level instructions since they feature different input lengths, levels of detail, and word choices. Our results show there is a significant gap between the performance of high-level instructions and human instructions, showing that current models are not capable of handling the increased complexity in language. Among all the results, models perform significantly worse on Stack2 and Pass2 with human instructions. This discrepancy is likely due to the fact that the orders of objects vary in human instructions for these tasks (e.g. Place the red cube under the white cube" vs "Place the white cube on top of the red cube"), as demonstrated in \Cref{tab:instruction}, which poses greater challenges in language understanding. In addition, for tasks requiring tool use, human instructions do not always involve the tools, e.g. one instruction for poke is "push the red block toward the other robot and put it on the blue circle". In these cases, the model often fails to predict the correct object to manipulate.

\subsubsection{Further Analysis}
To provide more insight into the factors affecting task performance, we further show a breakdown of performance under different settings, including robot types and the number of distractors in the scene. We observe that in general, the task performance decreases with an increasing number of distractor objects (\Cref{tab:distractors}). As for the robot types, the collaborations among two UR10s exhibit significantly higher performance using the hierarchical planning model given high-level instructions (\Cref{tab:robot_types}). This is probably due to the fact that UR5s have a smaller reachable space compared to UR10s, making it more critical to accurately predict the reachable workspace of each robot.


\section{Conclusion, Limitations and Future Work}
In this paper, we presented \benchmark, the first public benchmark for language-conditioned multi-robot tabletop manipulation. \benchmark combines the problems of language grounding, task planning, task allocation, tool use, capability estimation, long-horizon manipulation, and multi-modal scene understanding. All these subproblems of \benchmark pose significant challenges for existing algorithms.  We plan to open-source the simulation environment, the generated dataset, the baseline models, and other tools used during the project's development stage, and we hope our benchmark can inspire new methods in these areas. In this work, we assume tasks can be easily decomposed into independent sub-goals and we restrict the object manipulation problem to SE2. Extending it to SE3 with a more diverse set of tasks and objects will be an important future direction. In our setting robots have different capabilities if they have different reachable spaces. Of course, robots can differ in multiple other ways such as payload, gripper types, etc. We use instructions generated by templates, which are not as diverse as human natural language instructions. Leveraging LLMs to produce more diverse instructions can be a promising way to increase linguistic diversity. This work did not specifically focus on bimanual robot control, and one important future direction is to explore challenges inherent to bimanual control including dual-arm coordination and collision avoidance with real robot experiments. However, this study is the first of its kind to study vision-based language-conditioned multi-robot collaboration and our simplifications serve as a reasonable starting point for research in this area. 

\bibliographystyle{IEEEtran}
\bibliography{IEEEexample}

\end{document}